\documentclass[manuscript,authorversion,nonacm]{acmart}

\AtBeginDocument{%
  \providecommand\BibTeX{{%
    \normalfont B\kern-0.5em{\scshape i\kern-0.25em b}\kern-0.8em\TeX}}}

\acmJournal{JACM}
\acmVolume{37}
\acmNumber{4}
\acmArticle{111}
\acmMonth{8}

\acmPrice{15.00}
\acmISBN{978-1-4503-XXXX-X/18/06}

\usepackage{cleveref}
\usepackage{adjustbox}

\begin{document}

\title{Towards Avoiding the Data Mess: Industry Insights from Data Mesh Implementations}

\author{Jan Bode}
\email{jan.bode1@ibm.com}
\authornote{This manuscript does not represent an official IBM statement.}
\orcid{to be entered}
\affiliation{%
  \institution{IBM}
  \country{Germany}
}

\author{Niklas Kühl}
\email{kuehl@kit.edu}
\orcid{0000-0001-6750-0876}
\affiliation{%
  \institution{KIT}
  \country{Germany}
  }

\author{Dominik Kreuzberger}
\email{dominik.kreuzberger@ibm.com}
\orcid{to be entered}
\authornotemark[1]
\affiliation{%
  \institution{IBM}
  \country{Germany}
}

\author{Sebastian Hirschl}
\email{sebastian.hirschl@de.ibm.com}
\orcid{to be entered}
\authornotemark[1]
\affiliation{%
  \institution{IBM}
  \country{Germany.}
}

\renewcommand{\shortauthors}{Bode and Kühl, et al.}


\begin{abstract}
With the increasing importance of data and artificial intelligence, organizations strive to become more data-driven. However, current data architectures are not necessarily designed to keep up with the scale and scope of data and analytics use cases. In fact, existing architectures often fail to deliver the promised value associated with them. Data mesh is a socio-technical, decentralized, distributed concept for enterprise data management. As the concept of data mesh is still novel, it lacks empirical insights from the field. Specifically, an understanding of the motivational factors for introducing data mesh, the associated challenges, implementation strategies, its business impact, and potential archetypes is missing. To address this gap, we conduct 15 semi-structured interviews with industry experts. Our results show, among other insights, that organizations have difficulties with the transition toward federated data governance associated with the data mesh concept, the shift of responsibility for the development, provision, and maintenance of data products, and the comprehension of the overall concept. In our work, we derive multiple implementation strategies and suggest organizations introduce a cross-domain steering unit, observe the data product usage, create quick wins in the early phases, and favor small dedicated teams that prioritize data products. Whereas we acknowledge that organizations need to apply implementation strategies according to their individual needs, we also deduct two archetypes that provide suggestions in more detail. Our findings synthesize insights from industry experts and provide researchers and professionals with preliminary guidelines for the successful adoption of data mesh.
\end{abstract}



\maketitle

\section{Introduction}
\label{introduction}
As the volume of data continues to grow, organizations are striving to become more data-driven in order to outperform the competition \cite{Galer2022WhatsBusiness}. \footnotetext{VERSION 1.0 as of 31/01/2024} The International Data Corporation (IDC) forecasts the amount of data to more than double in the years 2022 - 2026 with private organizations leading the growth \cite{Reinsel2018TheCore}.

However, in the rapidly evolving landscape of data management, the limitations of traditional centralized data architectures based on data warehouses and data lakes are becoming apparent. These systems struggle to keep pace with the increasing volume and variety of data, posing significant challenges for central IT departments \cite{Dehghani2022DataScale, Machado2021DataArchitectures, Dehghani2019HowMesh}. 

Data produced in an increasingly decentralized manner and at a higher volume strains the capacity of these departments and leads to prolonged response times for data requests \cite{Vestues2022AgileStudy}. This delay is a critical bottleneck impacting data consumers' accessibility to relevant data and decreasing the overall agility and responsiveness of the organization in a data-centric environment.

Complicating matters further, the growing variety of data introduces an additional layer of complexity. Central IT, often lacks the understanding of specific domains required to manage this variety effectively. This gap in domain-specific knowledge prevents the accurate and efficient handling of data, leading to mismatches between data provision and the actual needs of different organizational units \cite{Butte2022}.

Moreover, the centralized approach to data management raises significant concerns regarding data ownership within the broader field of data governance \cite{Dehghani2022DataScale}. In the absence of clear data ownership, responsibility for data quality and maintenance becomes ambiguous, leading to potential issues in data integrity and quality.

These challenges collectively obstruct seamless access to high-quality data, diminish data integrity, and inflate the time to market and value realization for data-driven initiatives \cite{Butte2022}. Consequently, the potential scale and effectiveness of data and artificial intelligence (AI) applications are limited, hindering the organization's evolution into a fully data-driven entity.

To tackle the shortcomings of centralized data management approaches, Dehghani \cite{Dehghani2019HowMesh} postulates data mesh as a new socio-technical data concept. 
This concept leverages the distribution of data ownership across specific domains, directly tackling the inefficiencies of central IT overload and enhancing the quality of data management with domain-specific insights. It introduces a paradigm where data is treated as products, emphasizing its quality, usability, and lifecycle, thereby addressing concerns around data integrity and ownership clarity. Empowering users with a self-serve platform significantly reduces delays in data access, aiming to boost organizational agility. Furthermore, federated data governance ensures data standards are maintained without stifling innovation, aligning with the unique needs of each domain. This strategic alignment of principles aims to mitigate current data management challenges and enable data democratization\footnote{Data democratization refers to universal data access for all employees across an organization \cite{Awasthi2020ADemocratization}.}.

Whereas Dehghani's \cite{Dehghani2022DataScale} work lays a foundation for decentralized data architectures, empirical research on this topic is still scarce \cite[c.f.]{AraujoMachado2022AdvancingImplementations, Vestues2022AgileStudy, Machado2021DataArchitectures}. Recent publications provide insights regarding technological data mesh architectures \cite{AraujoMachado2022AdvancingImplementations, Butte2022}, privacy challenges \cite{Podlesny2022CoK:Meshes}, the overall concept \cite{Priebe2021FindingArchitectures, Machado2021DataArchitectures}, or describe single case studies \cite{Vestues2022AgileStudy, Joshi2021DataStudy}. 

However, no scientific work empirically investigates how industry experts apply the high-level concept of data mesh across multiple industries. This is problematic because substantial changes in enterprise data management are a huge organizational effort. Gaining insights into other organizations' experiences with implementation---identifying the challenges they encountered and the strategies they deployed---is instrumental to avoiding costly pitfalls and preventing a data mess. Additionally, it's crucial to explore the motivational factors driving the adoption of data mesh, as this can inform organizations about its feasibility and suitability for their specific needs. An examination of the overall impact of adopting a data mesh is also essential, as it informs on the broader implications of such a shift. Further, utilizing archetypes can offer more tailored advice, recognizing that different types of organizations have distinct requirements and challenges.
Consequently, we formulate the following research question:

\textbf{RQ:}\textit{What preliminary insights can we derive for data mesh adoptions along the categories of motivational factors, typical challenges, implementation strategies, real-world impact, and resulting archetypes?}

To address this question, we first conduct an exploratory literature review followed by a series of semi-structured expert interviews across multiple industries. 
This work contributes preliminary insights beneficial for professionals who are in the process or plan to adopt a data mesh as an architectural data governance paradigm. We emphasize, that the contribution of this work is theoretical and conceptual---addressing foremost aspects related to the organizational and social dimension of data mesh. For technical implementations, we refer to publicly available guides \cite[c.f.]{aws} and recent publications \cite[c.f.]{AraujoMachado2022AdvancingImplementations}.

We structure the remaining article as follows. \Cref{foundations} introduces the theoretical foundations and discusses related work. Subsequently, \Cref{methodology} describes the methodology, whereas \Cref{results} presents the results from the analysis of the expert interviews. We highlight our contributions and point out future work and limitations in \Cref{discussion}. Finally, we conclude our work in  \Cref{sec:conlusion}.

\section{Foundations \& Related Work}
\label{foundations}
Organizations have to continuously re-think and adopt their data strategies, architectures, and management systems to create value from an ever-increasing amount of data to stay competitive in the field \cite{Sagiroglu2013BigReview}. In the past, various terminologies have emerged around the related concepts, including but not limited to terms such as ``data warehouse'', ``data lake'', and, more recently ``data lakehouse'', ``data mesh'', and ``data fabric''. In this foundational section, we clarify these terms, their core concepts, and interconnections.

In general, data warehouses and data lakes focus on data management, whereas data lakehouses, data mesh, and data fabric broadly refer to data architectures \cite{groger2021there}. Data management systems and architectures differ in the level of abstraction. For instance, a data architecture may include and orchestrate multiple data management systems \cite{orenga2018framework}. 

Data warehouses are specific databases that include structured data from multiple sources and mainly serve as central storage for processed data \cite{Vaisman2014DataConcepts}. They traditionally store data for business intelligence and reporting purposes and thus refrain from the storage of data for future exploration \cite{inmon2005building}.
In comparison, data lakes are able to ingest data at greater speeds and store higher volumes, as well as different types of data \cite{couto2019mapping}. In contrast to data warehouses, data lakes additionally store raw data for future exploration and possible business activities. As such, they are of great importance for machine learning (ML) applications. 
The recently emerged architecture combines the flexible storage of data lakes with the analytical structure of data warehouses, offering a scalable solution for managing and analyzing diverse data types. This hybrid model enhances data accessibility and analytics, addressing the evolving needs of big data management \cite{Harby2022}.
In accordance with related literature, we summarize and define the three terms as follows:

\textbf{Definition 1: Data warehouse.} 
\textit{A data warehouse is a subject-oriented, integrated, nonvolatile, and time-variant collection of data in support of management’s decisions} \cite{inmon2005building}.

\textbf{Definition 2: Data lake.} 
\textit{Data lake is a central repository system for storage, processing, and analysis of unstructured, semi-structured, or structured raw data in its original format} \cite[based on]{couto2019mapping}.

\textbf{Definition 3: Data lakehouse.} 
\textit{A data lakehouse is a data management system based on low-cost and directly-accessible storage (of structured, semi-structured, and unstructured data) that further provides traditional analytical DBMS management and performance features} \cite[based on]{armbrust2021lakehouse}.

A related but different term is data fabric. Data fabric is a technical architecture that brings together heterogeneous data that spans across multiple data sources; it allows organizations to monitor and manage data regardless of the location, considering appropriate data governance and data cataloging \cite{Li2022AGraph}. 
As such, its' primary focus lies in the integration of multiple data management systems, including data warehouses, data lakes, or data lakehouses.
To provide access for users throughout the organization, data fabric uses rich metadata \cite{stoyanovich2022responsible} and a data virtualization layer \cite{Li2022AGraph}. The use of metadata is crucial not only for accessing, discovering, and understanding of data, but also for automating data integration, engineering, and data governance activities. This includes centralized management of data access, privacy, and compliance-related topics. 

\textbf{Definition 4: Data fabric.} 
\textit{A data fabric is an emerging data management design for attaining flexible, reusable, and augmented data integration pipelines, services, and semantics} \cite{datafabricgartner}.

\begin{figure}
    \centering
    \adjustbox{width=1\textwidth}{
    \includegraphics{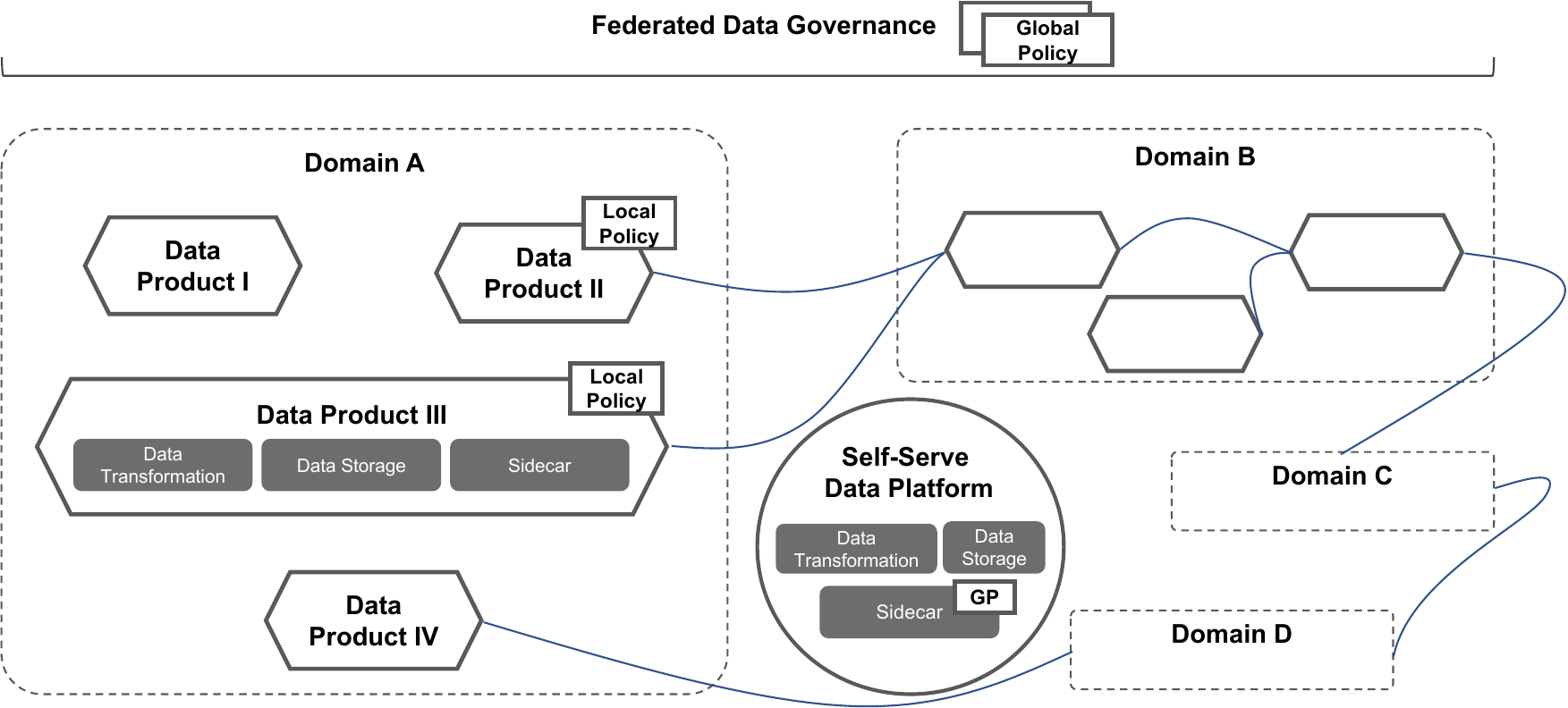}}
    \caption{Conceptual overview of a data mesh based on the four key principles: 1) domain-oriented decentralized data ownership, 2) data as a product, 3) self-serve data platform, and 4) federated data governance. The figure shows different levels of granularity (high on the left and low on the right).}
    \label{fig:architecture}
\end{figure}

In contrast, data mesh is a socio-technical concept, including architectural aspects. It further incorporates social and organizational aspects like decentralization and ownership. Similar to data fabric, a data mesh usually consists of multiple data management systems enhanced by an integration and governance layer and paired with a decentralized organizational structure. According to reference \cite{Dehghani2022DataScale} data mesh consists of four main principles that allow organizations to manage data at scale.
First, \textit{domain-oriented decentralized data ownership}: individual domains own the data they produce and leverage their domain knowledge to improve data quality. We define domains as organization-specific delineations of the relevant competitive boundaries of the organizations \cite{Sidhu2000BusinessPerformance}. Consequently, domain knowledge means that someone has expertise in a specific field or area gained through experience, education, or training.
For instance, the production department acts as a domain that owns all production-related data as they have the greatest expertise and are able to understand complex technical relationships reflected in the data.
Second, \textit{data as a product}: data are treated as products with end-to-end responsibility. Data products are provided\footnote{The provision of data products includes the initial development, making the data accessible, and maintaining the data product. In the following, we only refer to \textit{provision}.} including metadata, accessibility options e.g., APIs, and the actual data. It is the equivalent of a software product that also requires additional services, such as security updates or manuals. Furthermore, data products adhere to the following usability characteristics: discoverable, addressable, understandable, trustworthy, accessible, interoperable, valuable, and secure \cite{Dehghani2022DataScale}.
The third principle, \textit{self-serve data platform}, describes a dedicated data platform that provides high-level abstraction infrastructure for the domains; enabling domains to work highly autonomously. This is crucial for domains to avoid the replication of technical efforts and instead focus on the creation of high-quality data products.
The fourth and last principle: \textit{federated data governance} defines the governance structure for data products. Domain data product owners and relevant stakeholders collaboratively decide on common standards and policies---to be enforced automatically within each domain---to ensure the interoperability of data products.
This is of utmost importance because data products create the greatest value when combined.
In combination, the four principles allow organizations to overcome the limitations of centralized data architectures and enable organizations to become more data-driven \cite{Dehghani2022DataScale}. A depiction of the resulting architectural concept can be found in \hyperref[fig:architecture]{Figure 1}.

The adoption of a data mesh involves three key phases: \textit{exploration and bootstrapping}, \textit{expand and scale}, and \textit{extract and sustain} \cite{Dehghani2022DataScale}. Initially, select domains act both as data providers and consumers, establishing foundational practices and integrating data into aligned products. As the mesh grows in the expand and scale phase, an increasing number of domains join, standardizing technical and organizational patterns to enable rapid scaling and integration of legacy systems. Finally, in the extract and sustain phase, domains achieve autonomous data ownership, focusing on optimizing and refining data product delivery and usage, ultimately leading to a mature, cohesive data ecosystem. Each phase builds upon the previous one to enhance scalability and integration across the organization.

We want to stress that data mesh has architectural aspects as outlined before, but at the core is a socio-technical concept. Thus, adapting from reference \cite{Dehghani2022DataScale}, we define data mesh as follows.

\textbf{Definition 5: Data mesh.} 
\textit{Data mesh is a socio-technical, decentralized, distributed concept for enterprise data management.}

Due to the prevailing confusion surrounding the differentiation of data mesh and data fabric \cite[c.f.]{Priebe2021FindingArchitectures}, we highlight similarities and distinctions in the following.

Both data mesh and data fabric share common objectives related to data accessibility and interoperability, with the greater goal of enabling data-driven organizations. They both aim to dismantle data silos, enabling the seamless flow of data throughout the organization, leveraging self-service access to ensure data availability to those who require it. Further, both data mesh and data fabric put a strong emphasis on scalability to accommodate increasing volumes of data.

However, they greatly differ in their organizational approach. Data mesh advocates for decentralization of data ownership---within a decentralized organization---whereas the concept of ownership is not clearly defined for data fabric. As such, data mesh encompasses more social and organizational aspects, whereas data fabric puts a strong focus on technology solutions.

In conclusion, data warehouses offer a solution for storing data and supporting business intelligence applications, whereas data lakes serve as a repository for storing data and facilitating ML and data science projects. Data lakehouses merge these functionalities, providing both storage and advanced analytical capabilities, encompassing business intelligence, ML, and data science.

Data fabric and data mesh both (usually) encompass and orchestrate multiple data management systems and focus on integration, governance, and accessibility. Further, data mesh additionally encompasses organizational and social components, whereas data fabric has a stronger technological emphasis.

With these foundations at hand, we review recent literature addressing the term data mesh and its related concepts.

Reference \cite{Dehghani2019HowMesh} first introduced the term data mesh in May 2019. Since then, multiple organizations shared knowledge, PoVs, and their experience in the field of data mesh 
\cite[c.f.]{Thoughtworks2023DataThoughtworks,Whyte2022DataPlatform}.
However, peer-reviewed publications that contribute to the knowledge base of the novel topic of ``data mesh'' are still scarce. 
Recent peer-reviewed publications explain the term data mesh \cite{Priebe2021FindingArchitectures, Machado2021DataArchitectures} and introduce descriptions of multiple data mesh adoptions. However, those descriptions lack detail and only aim to explain or illustrate the concept. 
Two recent case studies \cite{Vestues2022AgileStudy, Joshi2021DataStudy} provide in-depth insights into data mesh adoptions in the \textit{Norwegian Labor and Welfare Administration} and the \textit{Saxo Bank} respectively. Both studies contribute valuable knowledge but are tailored to sole case studies. 

In summary, there is existing work on the topic of data mesh, however, concerning the RQ at hand which focuses on motivational factors, challenges, and implementation strategies---real-world industry insights are missing. Thus, we choose a qualitative interview-based study with experts from the field as the means of empirical research to explore this novel area.

\section{Methodology}
\label{methodology}
In order to gain a comprehensive overview of the motivational factors, challenges, and implementation strategies of data mesh adoptions, 15 semi-structured
expert interviews are conducted between November 2022 and January 2023. 
The method of semi-structured interviews is chosen for its ability to balance between the structured nature of closed questions and the flexibility of open-ended questions. This balance is crucial for exploring complex and novel topics, such as data mesh, while allowing new ideas and themes to emerge during the interview process \cite{semistructured}. 
Following the approach of reference \cite{Helfferich2011DieDaten}, an interview guideline is
used, structuring the interviews with regard to the outlined topics. 
The guideline was initially developed to reflect the core themes of our research question. Through pilot testing, we assessed the effectiveness of our approach, making refinements to the guideline. Throughout the interview process, we further tailored the guideline to include topics raised by the interviewees to ensure a comprehensive exploration of the subject. The main version of the interview guideline was established after the fourth interview.
We use a purposive sampling method \cite{Etikan2016ComparisonSampling} to interview partners from diverse industries to comprehensively address data mesh properties by incorporating a range of perspectives and applications. 
More specifically, we utilize expert sampling \cite{Etikan2016ComparisonSampling} by identifying experts based on their LinkedIn job titles and activities---most notably LinkedIn posts and comments. Additionally, we directly reach out to key stakeholders via LinkedIn who played a pivotal role in publicly available data mesh success stories. 

To qualify for inclusion in the interview process, candidates were required to have at least one year of experience in the domain of data mesh, along with five years of experience in the fields of data and AI. In one instance, an interviewee had only six months of experience in data mesh; however, this participant was included due to their extensive background in the closely related area of distributed data architectures. While one year may seem a minor inclusion requirement, it represents considerable expertise in the emerging domain of data mesh, which was only introduced in 2019\footnote{One participant stated five years of experience; the provided rationale was \textit{experience in the domain}, previous to the introduction of the term data mesh.}.

\begin{figure}
    \centering
    \adjustbox{width=0.7\textwidth}{
        \includegraphics{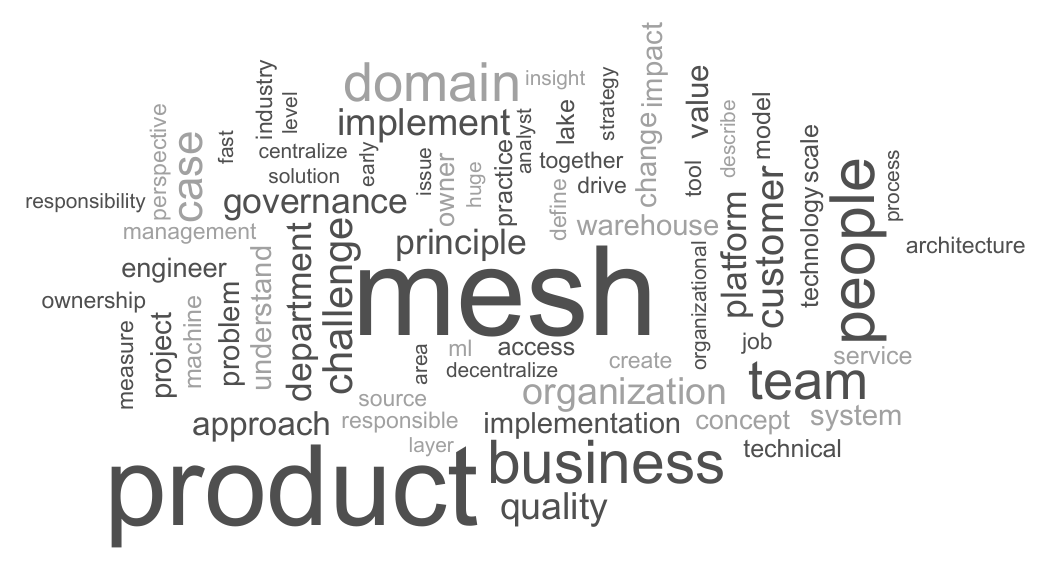}
    }
    \caption{World-cloud of main concepts discussed during the interviewees. Stop-words were excluded and terms lemmatized. Concepts with a larger type-size represent a higher relevance across interviews based on linear scaling. For reference, \textit{mesh} was found 570 times, whereas \textit{ml} was found 51 times. The term \textit{data} was excluded because its high frequency of 2447 would have skewed the visualization of other concepts.}
    \label{fig:cloud}
\end{figure}

Our study encompasses representation from companies of varying sizes and levels of experience with the topic. An overview of the interviewees and their characteristics are depicted in \Cref{tab:interviews}. Further \hyperref[fig:cloud]{Figure 2} provides and overview of the main concepts discussed during the interviews. 

\begin{table*}[htbp]
  \caption{Overview of the interviewees, their acronyms, title, experience, industry, company size, and length of interviews.}
  \label{tab:interviews}
\adjustbox{width=1\textwidth}{
  \begin{tabular}{ccccccc}
    \toprule
    Name & Job Title & Experience & Experience & Industry & Company & Length\\
    & & Data \& AI (y) & Data Mesh (y) & & Size (\# emps.) & (mm:ss) \\
    \midrule
    $A$ & Solution Architect & 9 & 2 & Sportswear & 57.000 & 58:52\\
    $B$ & Director Data of \& Analytics & 15 & 2 & E-commerce & 17.000 & 23:11 \\
    $\Gamma$ & Senior Consultant & 5 & 1 & Consulting & 150 & 41:38 \\
    $\Delta$ & Director of Data \& Analytics & 10 & 3 & Consulting & 300.000 & 36:52 \\
    $E$ & Data Analyst & 6 & 3 & E-commerce & 17.000 & 50:29 \\
    $Z$ & Technical Specialist Data \& AI & 30 & 3 & IT & 290.000 & 51:21 \\
    $H$ & Lead Engineer & 7 & 3 & Automotive & 120.000 & 39:07 \\
    $\Theta$ & Corp. Transformation Architect & 20 & 3 & Software & 110.000 & 42:11 \\
    $I$ & Director of Engineering & 10 & 2 & Food delivery & 5.000 & 45:23 \\
    $K$ & Solution Advisor Chief Expert & 40 & 3 & Software & 110.000 & 46:22 \\
    $\Lambda$ & Head of Diagnostics Data Office & 20 & 3 & Healthcare & 100.000 & 45:59 \\
    $M$ & Senior Manager Solution Architect & 8 & 5 & Software & 4.000 & 23:42 \\
    $N$ & Technical Lead Data \& AI DACH & 6 & 0.5 & IT & 290.000 & 25:45 \\
    $\Xi$ & Senior Manager Data \& AI Strategy & 12 & 2 & Consulting & 415.000 & 49:39 \\
    $O$ & Chief Data Architect & 25 & 3 & Automotive & 37.000 & 52:55 \\
  \bottomrule
\end{tabular}
}
\end{table*}

Due to the COVID pandemic, all interviews were conducted via video calls and recorded with consent. 
Moreover, this shift to a digital format allowed us to engage with a wider array of experts spread across various locations, facilitated the recording processes, and offered greater scheduling flexibility to accommodate the experts' commitments.

Following the interview, recordings were transcribed for qualitative content analysis using open, axial, and selective coding \cite{Krippendorff2018ContentMethodology, williams2019art, lincoln1985naturalistic,flick2022introduction} and MAXQDA2020 \cite{Radiker2020MAXQDAMAXQDA}.

Each interview underwent an initial independent analysis using open coding, which was subsequently complemented by axial coding to condense and allow deductions across interviews. Finally, axial codes were sorted into themes using selective coding. This process was repeated 15 times, with each interview representing one iteration.
This approach facilitated the extraction of pivotal insights and also allowed their incorporation into a broader framework \cite{williams2019art}. 

\begin{figure}
    \centering
    \adjustbox{width=1\textwidth}{
    \includegraphics{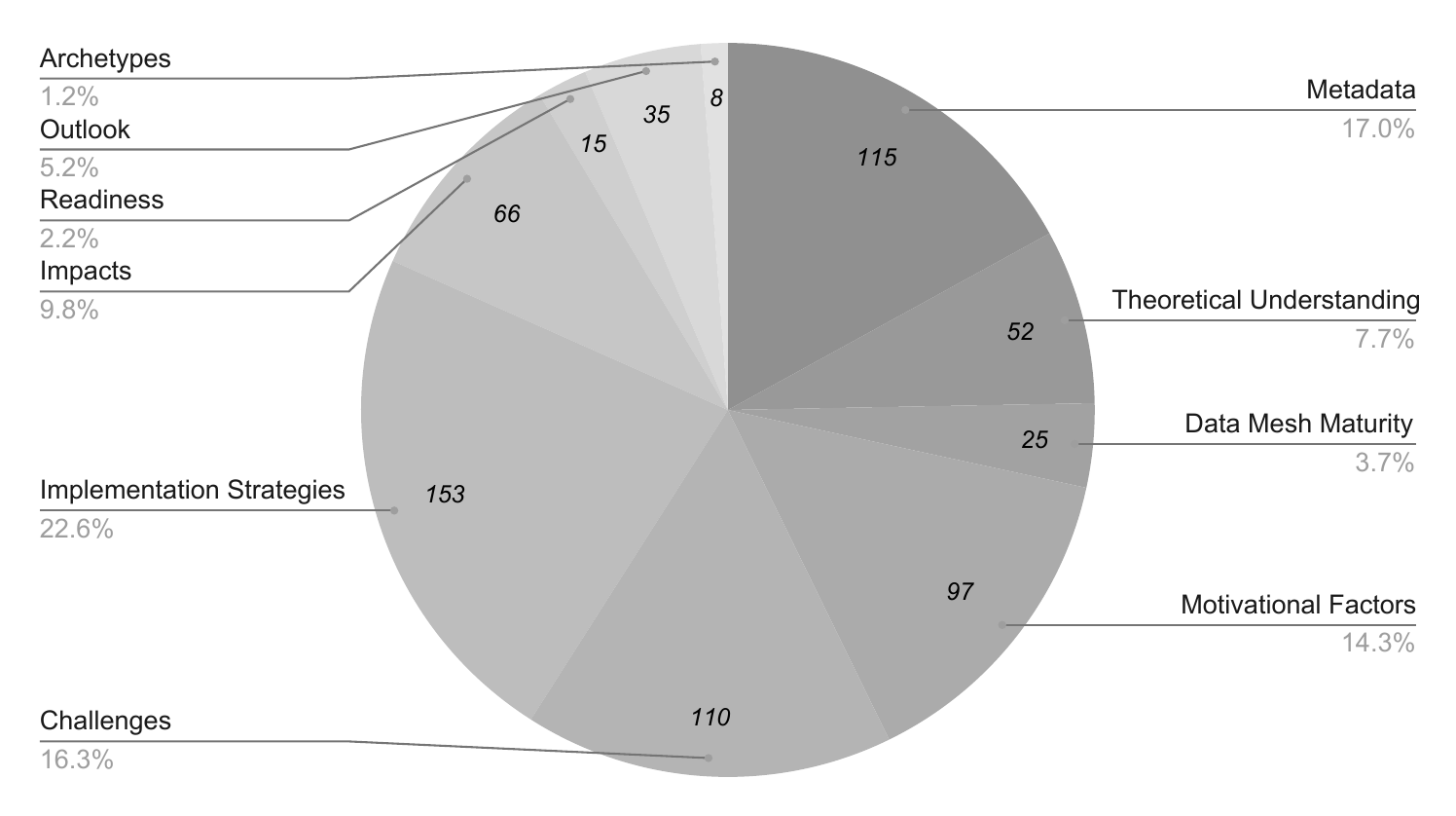}}
    \caption{Pie-chart of interview themes. Themes are sorted clockwise according to the interview guideline. 39 codes of the archive theme are omitted.}
    \label{fig:themes}
\end{figure}

More specifically, in each iteration we first conducted a thorough review of the automatically produced transcripts to guarantee the accuracy of the content. Text segments were subsequently, paraphrased and condensed to provide a clearer overview \cite{flick2022introduction}. Following that, we established the first coding iteration on an interview-level, based on the paraphrased segments. Thereafter, codes were reviewed to guarantee they accurately reflected the content of the interviews.
In the second stage, interview-level codes were integrated into the overall framework. For that purpose, we first sorted codes into the following main themes: \textit{theoretical understanding}, \textit{case description}, \textit{motivational factors}, \textit{challenges}, \textit{implementation strategies}, \textit{impacts}, \textit{readiness}, \textit{outlook}, and \textit{archetypes} to allow deductions with regard to the established research question. \hyperref[fig:themes]{Figure 3} visualizes the main themes with their relative and absolute distributions in a pie-chart.

\begin{figure}
    \centering
    \adjustbox{width=0.8\textwidth}{
        \includegraphics{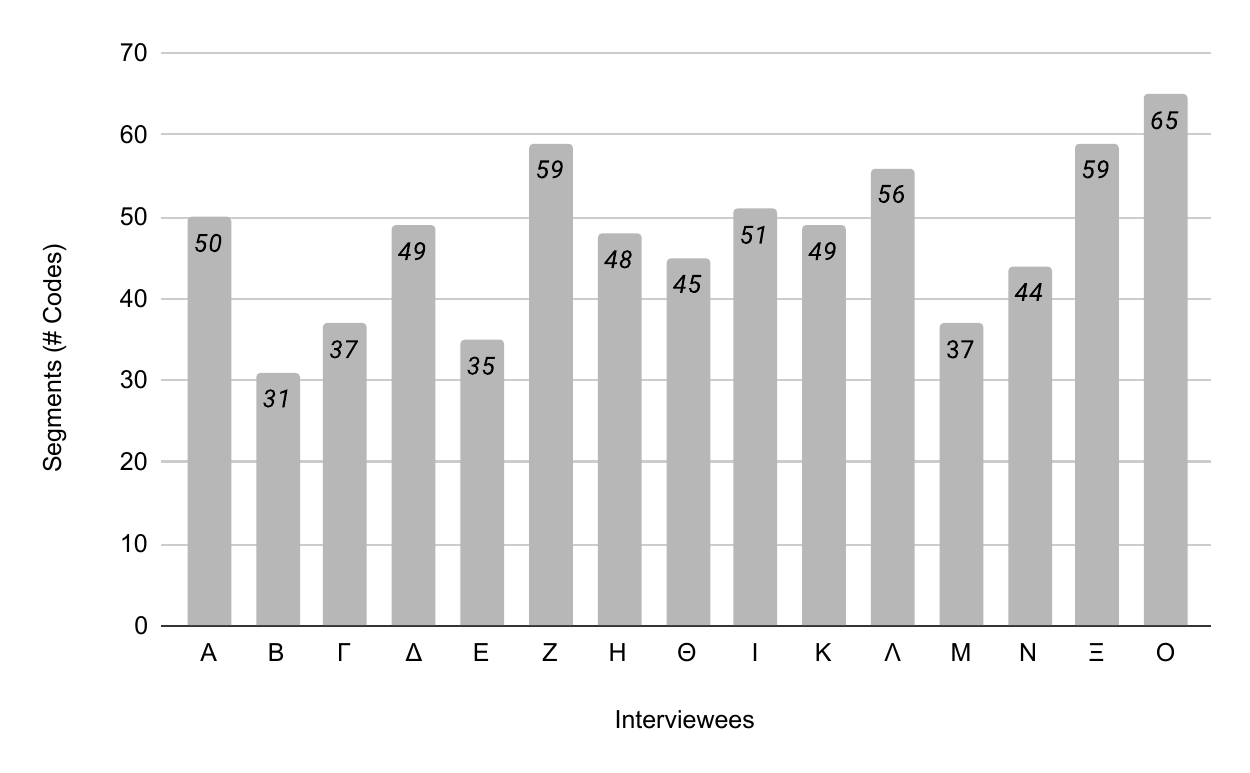}
    }
    \caption{Bar-chart of coded segments for each interviewee. Number of coded segments on the y-axis; interviewees as defined in \Cref{tab:interviews} on the x-axis.}
    \label{fig:interviewees_codes}
\end{figure}

In each iteration, we created and adjusted sub-codes, within each theme to cluster similar statements from multiple interviewees into axial codes \cite{lincoln1985naturalistic}.
After the completion of the axial coding, we refined each sub-group to obtain selective themes represented in the motivational factors, challenges, implementation strategies, impacts, and archetypes presented in the following section. 
In total, we derive 717 (sub)-codes, across 15 interviewees resulting from 48 hours of coding work. \hyperref[fig:interviewees_codes]{Figure 4} displays the absolute number of coded segments across interviewees.

\section{Results}
\label{results}
In this section, we synthesize findings from the interviews. We provide insights regarding interviewees' \textit{theoretical understanding}, their \textit{motivational factors} for the adoption of a data mesh, \textit{challenges} they face, and \textit{implementation strategies} they derive. Furthermore, we focus on the \textit{impact} interviewees observe. Finally, we present two \textit{archetypes} of organizations that adopt the data mesh concept. 
We focus only on aspects that are highly relevant for data mesh applications. However, some \textit{challenges} and \textit{implementation strategies} are not unique to data mesh but rather relate to the overall topic of change management and technology introduction \cite[c.f.]{By2007OrganisationalReview, Oakland2007SuccessfulManagement}.

\subsection{Theoretical Understanding}
Interviewees generally support the main principles of data mesh introduced by Dehghani ($\Theta-M, O$) and emphasize the importance of the combination of all four main principles to adopt a successful data mesh. However, interviewees assign the highest value to the principles of domain-oriented decentralized data ownership and data as a product when asked to favor one over the other.
Furthermore, interviewees state the need for further guidelines that assist in the adoption and operation of a data mesh ($A, \Delta, E, \Theta, K$). $A$ states: ``the concept is described as highly theoretical, and I believe it is a challenge to transfer this theory into practice for many organizations'', whereas $E$ describes the main principles as ``too broad'' to adopt a data mesh out of the box---highlighting the importance to provide practice-oriented guidelines for professionals going forward.

\subsection{Motivational Factors}
Prior to the adoption of the data mesh concept, interviewees face a variety of problems and thus consider multiple motivational factors. 

\textbf{MF1: Reduce bottlenecks} ($A-E, H, \Theta, K-O$). The first factor combines the related concepts of bottlenecks, scalability, and speed and is consistent with the initial proposal of Dehghani \cite{Dehghani2022DataScale}.
Bottlenecks refer to a lack of capacity in the central data management system or the team responsible for the data provision and result in a backlog of data requests from multiple domains. As a consequence, domains are unable to scale data-driven use cases at the preferred rate, e.g., $M$ states ``in the food delivery space, they were not able to scale anymore because this big data warehouse in the middle was the bottleneck''.
The concept of speed has two dimensions. 
It describes the time the central team needs to provide requested data and the time to market for new products and services; $B$ highlights the importance: ``speed and delivery of shipping customer features, improving customer journeys, and scaling to more markets. We are hyper growth company. Speed matters when you're on a hyper-growth''.

\textbf{MF2: Leverage domain knowledge} ($A-Z, I, K, \Xi, O$). In centrally governed organizations IT departments provide data, however, they lack domain knowledge to ensure the task-specific quality of the data. Interviewees emphasize the importance of domain knowledge to provide data of a higher quality. $\Gamma$ emphasizes, ``the person didn't have the domain knowledge [...] therefore the results were bad'', and $E$ goes even further, stating ``we need to somehow give that capability of dealing with data back to the people who actually understand what the hell the use cases are.''

\textbf{MF3: Break down silos} ($A, \Gamma, E-I, \Lambda, M, \Xi, O$). Accessibility is of great importance for cross-domain data-driven use cases. However, silos within organizations prevent other domains to access the data. For example, the marketing department (domain A) may need to request sales (domain B) data to track the success of a regional marketing campaign. However, if sales fail to deliver the data, the marketing department loses out on valuable insights. Providing a real-world example $O$ states: ``We found out that nobody in the company has a full idea which kind of data exists and where the data is accessible.''

\textbf{MF4: Establish data ownership} ($I, \Lambda, N$). The fourth motivational factor highlights the significance of data ownership. It refers to the end-to-end responsibility of a data product in terms of quality and accessibility. This concept is comparable to traditional product ownership roles. $I$ describes the underlying problem in the following way: ``we had the problem that when we pushed the data to the central data. Data ownership was lost''. Further, $I$ relates the concept of stronger ownership with better data quality: ``I think it's very simple. The main reason we did this was to improve data quality. And the bet was that stronger data ownership would lead to better data quality.''

\textbf{MF5: Adopt modern architecture} ($Z, K, \Lambda$). Managers and employees within an organization are aware of an upcoming trend in data architectures---they see how organizations throughout the industry adopt a data mesh and ``don't want to miss the boat'' ($\Lambda$); ``some companies just heard from this hype and this is why they look into it'' ($K$). However, data mesh is not a universal approach. Hence this finding is very worrisome because it may push organizations or business units toward a data mesh adoption for the wrong reasons.

\textbf{MF6: Reduce redundancies} ($B, I, O$). Silos, poor communication, and lack of transparency can lead to redundant work in large organizations, including data-related areas like data preparation ($B, I$) or entire use cases ($O$). Various data consumers individually prepare data for their specific use cases but fail to share the results with the organization---``what we discovered in our research was that teams improve the data significantly, but don't share it back with the company'' ($B$). Similarly, $I$ points out: ``we're pushing all the way back to the data producer, who hopefully can solve the [data quality] problem instead of having to repeatedly deal with this issue further down the line''. As a consequence, redundant efforts across domains, divert resources away from value creation ($B$). Furthermore, interviewees from large, decentralized organizations complain that they spend significant resources on the provision of data for other departments---a task that data mesh aims to simplify ($I, O$).

After discussing motivational factors, it's crucial to recognize the challenges individuals may encounter when working with the data mesh concept. We now analyze obstacles that can hinder the progress of data mesh adoption.

\subsection{Challenges}
\label{subsec:challenges}
Organizations face multiple challenges throughout the adoption of data mesh. We mainly focus on challenges that are unique and of high relevance to data mesh-related topics. 

\textbf{C1: Federated data governance} ($A, \Delta-O$). We identify the shift from centralized toward federated data governance as the main challenge for professionals. Interviewees state that the federated approach introduces difficulties for activities and responsibilities previously managed centrally. Whereas they stress the importance of federated data governance to establish rules according to domain needs, interviewees highlight limitations regarding the automated execution; especially concerning security, regulatory, and privacy-related topics ($A, E-\Theta, N, O$). 
$O$ notes that employees within the domain are unaware of which data are protected and regulated.
$N$ warns that managers may ``end up with one foot in jail'' for non-compliance with data protection regulations. Furthermore, interviewees state the lack of a central unit results in insufficient prioritization of use cases and projects across domains ($\Delta, Z, I$). $I$ argues that decentralized ownership and federated governance result in a lack of observability, which in turn complicates proper prioritization. 

\textbf{C2: Responsibility shift} ($B, \Gamma, Z, \Theta, I-\Lambda, N$). The first main principle in reference \cite{Dehghani2022DataScale} states that ownership shifts from a central authority towards the domains. However, decentralized ownership within the domains includes end-to-end responsibility for the accessibility and quality of the data products. This new responsibility creates a number of sub-challenges. First, the data product owners perceive the task of providing data for other domains as extra work ($\Gamma, \Theta, K$)---``they had to fight with all the other items in the backlog and they didn't see any value in it'' ($\Gamma$). In addition, domains frequently receive no direct compensation for the data provision efforts that benefit other domains ($\Theta, K, N$). Furthermore, domains' business activity is usually not centered around the provision of data products---consequently, they deprioritize the task ($B, \Gamma, Z, I, \Lambda$). 

\textbf{C3: Metadata quality} ($E, \Theta, I$). The third challenge, as reported by interviewees, relates to the data product model. Data products store various data types and also offer metadata, aiding users without domain knowledge in data interpretation \cite{groger2021there, Dehghani2022DataScale}. However, interviewees point out a gap between the descriptions provider domains deliver and the information consumer domains need to correctly interpret the data. $E$ states that ``data modeling is the most crucial part in most companies and there is no standard yet on how to do this'' and subsequently elaborates that ``if you solve data modeling [...] you probably have the gold mine for data these days''. While the challenge of metadata quality is not unique to data mesh---``in former times, the problem already existed, but nobody really took care of it'' ($\Theta$)---it is complicated by unclear expectations regarding the data products and their use cases ($I$).

\textbf{C4: Comprehension} ($\Delta, Z, \Theta, \Lambda, X$). The fourth challenge relates to a lack of comprehension of the data mesh concept. Interviewees report that employees within the organizations simply use new terms without an organizational or technical change, e.g., to call all data sets data products ($\Delta$) or categorize a highly centralized data storage as a data mesh ($\Lambda$). $\Lambda$ further elaborates he/she is under the impression ``that the warehouse from yesterday is now called a mesh, just like a meeting today is called a sync''. Indicating a severe misinterpretation of the data mesh terminology among practitioners complicating efforts for data mesh initiatives to move forward.

Interviewees have pointed out additional challenges that, whereas not exclusive to data mesh, remain noteworthy. These challenges deserve attention because they can inform the development of specific implementation strategies for data mesh---neglecting them could potentially jeopardize the successful adoption of data mesh.

\textbf{C5: Resource limitations} ($A-\Delta, Z-\Theta, K-M, \Xi$). C5 describes the lack of financial, technical, and human resources---``It’s very resource intensive to build up a data mesh, and thus the data products. You need a lot of data engineers with ETL knowledge.'' ($A$). $\Theta$ further elaborates: ``everyone wants to get data-driven, but when it comes to the effort, companies are usually not eager to put more people or money into that''.

\textbf{C6: Acceptance issues and resistance}
($\Gamma, Z, \Theta, \Lambda, M, \Xi, O$). C6 relates to acceptance and resistance within organizations, for instances, $O$ states: ``Most of the time I was a somebody, let me say, a kinder-gardener. Yeah, just to tell the people, please change your mind. Please change your perspective. Please start looking at data from the perspective of the data''.

Both, C5 and C6 are well established in change management research, consequently, we do not discuss them any further but relate to the existing literature on this topic \cite{By2007OrganisationalReview, Oakland2007SuccessfulManagement,anastassiu2020resistance}.

\subsection{Implementation Strategies}
\begin{figure}
    \centering
    \adjustbox{width=1\textwidth}{
    \includegraphics{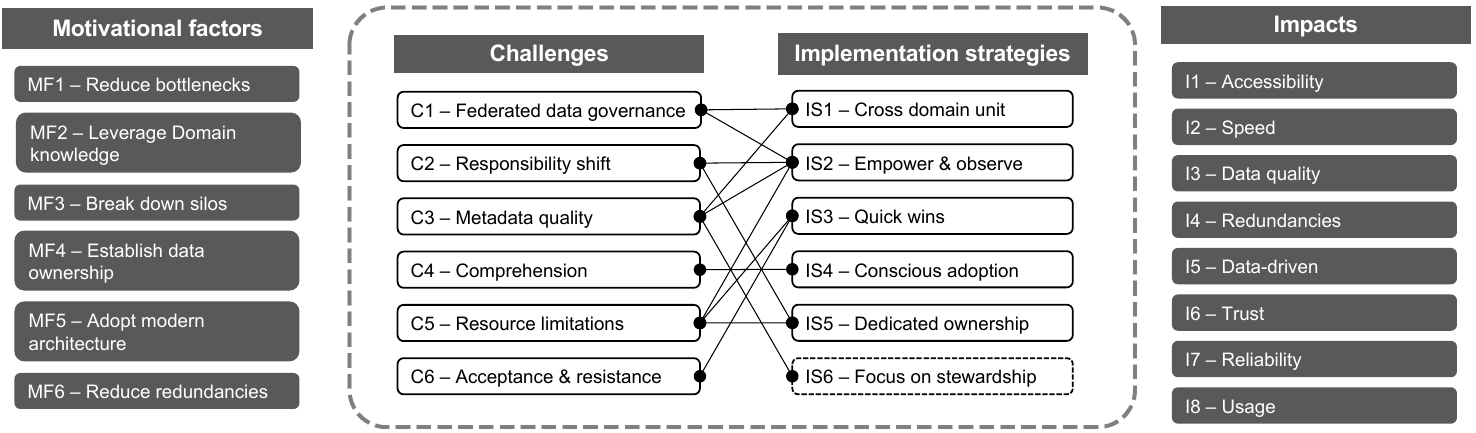}
    }
    
    \caption{Data Mesh-relationships between challenges and implementation strategies framed by motivational factors and impacts.}
    \label{fig:summary}
\end{figure}

In the interviews, we identify implementation strategies that support organizations to overcome the challenges we formulated in the previous section. We highlight relationships between challenges and implementation strategies to allow organizations to address challenges according to their needs. The implementation strategies are the main contribution of this work as they provide preliminary guidelines for professionals and researchers. In the following, we present implementation strategies and provide detailed descriptions. \hyperref[fig:summary]{Figure 5} on page \hyperref[fig:summary]{9} summarizes our central findings.

\textbf{IS1: Cross-domain unit} ($A, B, \Delta, Z-I, N$). Interviewees identify the transition towards federated data governance (C1) as the main challenge. To address this challenge, we argue that organizations should introduce a cross-domain steering unit responsible for strategic planning, use case prioritization ($A, \Delta, Z, H$), and the enforcement of specific governance rules---especially concerning security, regulatory, and privacy-related topics. This becomes necessary when full automation is not available yet. $\Delta$ further argues that human resource allocation from a central or a support team is essential in early phases, as it can help domains with strong use cases but little technical expertise. However, a cross-domain steering unit could also help to enforce high metadata standards (C3) ($\Delta, Z, H, \Theta, I$).
Furthermore, interviewees state that it can be beneficial to provide key data products centrally ($B, \Delta, N$). $B$ argues that central ownership in early phases provides a role model domains can follow for the provision of domain-owned data products. Furthermore, it ensures smooth operations of data that are of strategic value ($B, N$). $B$ elaborates: ``So one of the things that we do is we are taking 20 data sets that are of high, criticality. Datasets that we own centrally and turn into products. This way we act as a role model for other domains.'' 
Nevertheless, we stress that a cross-domain steering unit can only complement and support the federated governance structure and may grow obsolete with the increasing maturity of a data mesh, e.g., the fully automated policy execution that is centrally provided by the platform.
 
\textbf{IS2: Empower \& observe} ($B, \Delta, Z-K, M-O$). The second implementation strategy addresses the shift of responsibility (C2) and issues related to metadata quality (C3) and the transition towards federated data governance (C1). One of the key ideas of data mesh is to leverage domain knowledge. However, a central department needs to provide the technical infrastructure, support, templates, and guidelines to facilitate the provision of high-quality data products ($B, \Delta, Z-I, N$). This finding is in line with the third main principle of data mesh, the self-serve data platform highlighting the importance of empowering the domains. On multiple occasions during the interviews, interviewees stress the importance of observing the quality and usage of data products ($Z-K, M, \Xi, O$) pointing out two major benefits. First, usage data of data products can motivate data product owners within the domains because it visualizes the impact the data product has. This data can help domains highlight their importance within the organization and provide them with leverage to negotiate additional resources (C5). Furthermore, we suggest that organizations observe and automatically score the quality of data products, preferably by an independent unit. This score should include a metadata score, a data quality score, and availability metrics. Depending on the criticality of the data and data product quality standards, organizations should also consider a human-in-the-loop responsible for initial approval ($\Delta$).
$O$ argues that a data product quality score creates an incentive for data product owners to ensure a high quality of the metadata (C3). $\Delta$ further emphasizes, ``it doesn't need a lot of escalation [...] just by knowing it will be reported [...] they handle it very differently now''. Therefore, we suggest that a central steering unit tracks and ranks key data products to nudge data product owners to provide high-quality data products. 

\textbf{IS3: Quick wins} ($A-O$). Whereas the challenge of acceptance issues and push-back within an organization (C6) is well established, data mesh-specific implementation strategies can help to navigate organizational challenges.
We synthesize findings from interviewees to formulate a fine-grained multi-step process that guides professionals through the data mesh exploration \& bootstrap phase with respect to organizational challenges. 
First, data mesh initiatives should start ``where the budget is'' as financial resources are crucial for the success of the data mesh adoption ($Z$, C5). If central IT has the resources to kick-start the transformation towards a data mesh they should be responsible. If the organization is already highly decentralized, influential domains should lead the way. Either way, communication across multiple domains and stakeholder from central IT is crucial across the iterative transformation phase. In the next step, domains and central IT develop a concept that considers all stakeholders' interests to ensure future adoption, e.g., using surveys or interviews. Afterward, the main driver of the data mesh initiative should select multiple pilot data products. The pilot data products should exhibit the following characteristics: It should span across multiple use cases ($A, Z, \Theta, \Lambda$), be inexpensive ($A, \Delta, K-M$), be small ($A, \Lambda, M$) but impactful ($A, Z, I-\Lambda, N, O$), and allow for easy and quick wins ($A, H, \Lambda, \Xi, O$). Consequently, drivers of the data mesh initiative are able to present successful use cases to get approval from the top management and promote data mesh throughout the organization ($H, \Xi, O$). Simultaneously, the inexpensive and small approach guarantees acceptance for the initial pilot program. Moreover, educational initiatives and community-building efforts should support initial developments of data products. $\Delta$ emphasizes the significance of celebrating early achievements through events such as pitch nights, community gatherings, or hackathons to build momentum within the organization.

\textbf{IS4: Conscious adoption} ($B, \Delta, H, I-\Lambda, \Xi, O$).
Before organizations decide to adopt a data mesh, they should go through a thorough assessment of their current data management systems, architecture, its shortcomings, and the potential benefits of a data mesh. Moreover, they should consider the organizational structure and size. If the organization decides to move forward it is crucial to consciously and carefully introduce data mesh-related terminology. $\Lambda$ \& $O$ go as far as to suggest not to use data mesh-related terminology at all. $\Lambda$ \& $O$ argue that this creates a profound knowledge of the underlying concept and avoids hype-related misunderstandings (C4). Aiming to improve understanding, $I$ states: ``What I tend to do to convince people of this, I relate it to something that they know, which is typically an API. Basically, think of a data product as an API. You wouldn't release an API to the world without a team behind it and a well-defined definition of what it is''.

\textbf{IS5: Dedicated ownership} ($A, \Lambda, O$).
Responsibility shift (C2) highlights the challenge that domain employees perceive the provision of data products as extra work and do not prioritize the task. Since organizations have to work with limited resources (C5), managers need to carefully consider the number of employees and the capacities those employees can spend on the provision of a data product. Our interviews show that managers should favor a smaller team with people dedicated to data product provision tasks over a bigger team where commitment is lower for three main reasons. First, there is no prioritization conflict, because the main prioritization is the provision of data products. Moreover, a main prioritization shields the position from possible reorganization and restructuring efforts within the organization---``what happens when the reorganization comes along? [...] everything that is not your main job is eliminated. You do your main job or you get a new job'' ($\Lambda$). Last, dedicated data product owners can build stronger expertise, especially in the context of sensitive data and regulatory requirements ($O$).

\textbf{IS6: Focus on stewardship} ($O$). A data steward is a role within an organization or data product team that is responsible for the overall data management, such as rules for interaction with data \cite{otto2011morphology}. According to $O$, a data product steward owns the data products and ensures quality, fitness, and availability---including metadata quality (C3). Furthermore, a data product steward has complete knowledge of the existing use cases of a data product; this is a key extension of the data product owner role that focuses on the provision of data products for self-service. $O$ outlines a protocol wherein data consumers submit access requests for data products, providing a short explanation of their intended use. The steward is presented with multiple decision pathways: first, grant approval, second, grant approval and facilitate a connection with a related existing use case to explore potential synergies, third, reject the request due to regulatory or security constraints, or fourth reject due to the presence of an overlapping existing use case. For example, $O$ details an instance where a business unit intended to sell data already marketed by another part of the organization. Proceeding with this plan would have diminished revenues from the current sales and led to the duplication of organizational structures.
Consequently, stewards can leverage their knowledge to effectively reduce redundant work (MF6). However, professionals need to balance the advantages against the possible creation of bottlenecks (MF1) and silos (MF3) as previous approval may come at the expense of rapid development and self-service. In a data mesh environment, either the data product owner can take on these additional responsibilities or share them with another member of the data product team.

\subsection{Impact}
Out of 15 interviewees, ten share experiences from a transformation towards a data mesh in the late \textit{exploration \& bootstrap} phase ($\Gamma, \Delta, Z, \Theta, K-O$). Furthermore, eight interviewees share insights from the second, \textit{expand and scale} phase ($A, B, \Delta, E, H, I, \Xi, O$). Consequently, all interviewees are able to provide (preliminary) details on the impact of their data mesh adoptions.

\textbf{I1: Accessibility }($A-\Gamma, \Delta, Z-\Theta, \Lambda, N, O$).
The most prominent impact of data mesh adoptions that we have observed is the improvement in accessibility, which directly relates to the reduction of bottlenecks and silos (MF1, MF3); ``You can do your data analysis on your own without asking anyone. And if you find something to improve, you just do it on your own. And again, you have to ask nobody.'' ($\Gamma$). 

\textbf{I2: Speed }($\Gamma-E, H, I-\Lambda, \Xi$).
Furthermore, interviewees observe an increase in speed. As we describe in MF1, the reduction of bottlenecks relates to speed improvements along two dimensions. First, time to get access and second, time to market for new products and services. Interviewees observe improvements for both; ``a case that would have taken five months in the past, can be done now in three sprints of six weeks'' ($\Xi$).

\textbf{I3: Data quality }($A, \Gamma, \Delta, M-\Xi$).
Moreover, interviewees observe an increase in data quality due to clearly defined responsibilities; ``data quality is better as one data owner focuses on a specific data domain'' ($A$). $\Gamma$ reports the data to be ``much more up to date and less error-prone'', whereas $\Delta$ reasons that data owners have a stronger incentive due to increased accountability to deliver high-quality data products: ``if it gets tracked, it gets reported and of course, no one wants to be at the bottom of the list''.

\textbf{I4: Reduction of redundancies }($B, \Delta, H-I, M, O$).
``Develop once, use many times'' ($\Delta$); directly related to MF6, interviewees observe a successful reduction of redundancies. In particular, $B$ states that providing the most popular data sets in a prepared, clean format provides ``massive value because you have lots of engineering teams reusing those datasets''.
Additionally, $\Theta$ observes a more efficient extract, transform, load (ETL) process due to standardization: ``they implemented hundreds of data pipelines to hundreds of systems. Hundreds of times of effort. With data mesh, you simply provide a standardized API to the data marketplace''.
$O$ further mentions that domain knowledge can be transferred more easily.

\textbf{I5: Data-driven }($M-\Xi$).
Along with further impacts, we observe that organizations become more data-driven. This is in line with the overall strategy of organizations and according to interviewees correlates with a change in employees' mindset and a broad set of measures. 

\textbf{I6: Trust } ($I, K, M$).
Additionally, interviewees note an increase in data trust due to a higher domain knowledge (MF2) and data quality; ``suddenly, you have better trust in data because, behind the data, there are business people responsible for these data; the quality, the data, and the product managers.'' ($M$).

\textbf{I7: Reliability} ($A$). Moreover, $A$ reports higher reliability in comparison to the data warehouse concept as a consequence of improved accessibility and data provisioning; previously, the central database experienced crashes during peak demand periods when parallel queries led to system overloads and failures ``because of the distributed access we observed lower down-times of data access'' ($A$).

\textbf{I8: Usage} ($B, H, \Xi, O$).
Finally, interviewees report an ``explosive growth'' ($H$) in the number of use cases enabled by the improved data accessibility. More precisely, the lack of previous access barriers allows employees to develop new and iterate over use cases in a much shorter amount of time.  

Our observations indicate a generally positive impact from adopting data mesh, demonstrating that organizations can benefit even in the early stages of transition.

\subsection{Organizational Archetypes}
We acknowledge that the approach to an organization's data management systems and architecture is highly individual. Consequently, there is no one-size-fits-all approach and organizations need to weigh trade-offs and apply implementation strategies according to their needs. Nonetheless, in our findings, we identify two archetypes of organizations that share similar motivations, challenges, and implementation strategies. In this subsection, we describe both archetypes and highlight the most important findings.

\textbf{A1: Startup \& scaleup} ($B, \Gamma, \Delta, E, I, M, \Xi$). Startups \& scaleups are young organizations that operate in a dynamic, hyper-growth environment. They are early data mesh adopters and on average boast a higher readiness and maturity. Furthermore, startups \& scaleups usually consist of a younger workforce with a progressive organizational culture underpinned by a mature data culture where data and AI are integral to their product or service offerings. Due to the newness of the organization, startups \& scaleups usually lack legacy systems and are able to start architectural changes on a greenfield approach. Additionally, due to their recent establishment and a more limited range of products or services, these organizations often feature a more centralized structure. As a result, data mesh initiatives are typically driven top-down, supported strongly by top management.

Because of the hyper-growth environment, their focus is on speed and scalability (MF1). However, leveraging domain knowledge to improve data quality (MF2) and the reduction of redundancies (MF6) are also of great importance to improve ML capabilities and avoid redundant work in data preparation. MF1-2 and MF6 result in the following main challenges.

First, the dynamic environment and data-driven product and service offering typically require a high meta(data) quality (C3). Further, transitioning towards a federated organizational structure can be challenging (C1).
In contrast, acceptance and resistance to new initiatives (C6), as well as comprehension (C4), are typically less problematic due to flatter organizational hierarchies and the top-down approach. Neither are monetary resources (C5) typically the main challenge, as organizations favor speed and growth over profitability.

Given these settings, several implementation strategies are of great importance. The establishment of a cross-domain steering unit (IS1) is suggested to address standardization challenges and to provide clear guidance across domains. Although quick wins (IS3) are less critical in these top-down driven initiatives, empowering teams and observing outcomes (IS2) is crucial to creating the right incentives and ensuring the production of high-quality data products. Furthermore, dedicated ownership (IS5) is vital in maintaining clear responsibilities and accountability in a fast-paced environment, essential for the ongoing success and scalability of data mesh initiatives. IS6, focusing on stewardship, should be applied very carefully, as it has the potential to hinder the rapid development of new products or services. This is especially true since the duplication of whole use cases is less likely to occur in smaller organizations with lower hierarchies.

This approach allows startups and scaleups to effectively navigate their unique challenges and capitalize on their inherent agility and readiness for innovative data management practices.

\textbf{A2: Established organizations} ($A, \Gamma, \Delta, Z, H, \Theta, K, \Lambda, N, \Xi, O$). Established organizations have a long history and are large in terms of headcount and revenue. They consist of a senior workforce and possess a hierarchical structure with semi-autonomous units that cover multiple fields of business. Because of the company's history, established organizations frequently possess (multiple) legacy systems and more conservative organizational cultures. Regularly, they rely on centralized data storage concepts like data warehouses or data lakes.

The main motivational factors for established organizations to embrace data mesh are a reduction of bottlenecks and data silos for better accessibility (MF1, MF3), an improvement in (meta) data quality (MF2), clear data ownership (MF4), and a reduction of redundancies (MF6).

The primary obstacle they encounter pertains to organizational challenges. Organizational frameworks often tend to be inflexible, leading to issues related to the federated data governance (C1) and resistance (C6) within the organization. The rigid approach towards allocating financial resources (C5) hinders the required transformation for embracing data mesh. This is particularly noteworthy because quantifying the financial advantages of enhanced data quality and accessibility can be a complex endeavor.
Consequently, they do not provide good leverage for internal budget negotiations. Furthermore, expensive legacy systems make the business case for data mesh harder to justify. 

Given the outlined challenges, it is of great importance to create quick, effective, but inexpensive wins in the early stages of the adoption (IS3). With this approach data mesh initiatives can \textit{fly under radar} in the pilot stage and if successful leverage pilot data products or domains to promote widespread adoption and negotiate budget. 

Further, we suggest establishing a cross-domain steering unit (IS1) to consolidate multiple data mesh initiatives, enhancing overall data accessibility across the organization. Implementing dedicated ownership (IS5) and the empower and observe strategy (IS2) should also be considered, as these align well with the hierarchical structure of established organizations, ensuring clear responsibilities and fostering accountability within the data governance framework.

\section{Discussion}
\label{discussion}
In this section, we discuss results, highlight contributions, acknowledge limitations, and point out future work. 

\subsection{Summary \& Literature Embedding}
Data mesh is a socio-technical concept including architectural aspects that enables organizations to become more data-driven. However, the architectural concept of data mesh is novel and needs further scientific exploration---especially with regard to motivational factors, challenges, implementation strategies, impact, and archetypes. 
To derive industry insights we conducted 15 semi-structured expert interviews.

We find that interviewees' perception of data mesh and the associated motivational factors are generally in line with Dehghani's \cite{Dehghani2022DataScale} proposed concept. The interviewees formulate the need for faster, scalable solutions to reduce bottlenecks, leverage domain knowledge, and break down silos (MF1-3). In addition, the establishment of data ownership and the reduction of data redundancy are well documented (MF4, MF6). Furthermore, we find that the current hype around data mesh motivates stakeholders within organizations to adopt data mesh for the sake of adoption (MF5). However, data mesh is not a universal concept for data architectures and data mesh adoption should not become an end in itself. This insight highlights the need for clear fact-based discussions and communication. Reference \cite{Priebe2021FindingArchitectures} address this concern with a preliminary review of selected big data architectures.

In addition to motivational factors, we find several challenges that are partly reflected in current literature \cite[c.f.]{otto2022federated}. First, we identify the transition towards federated data governance as a main challenge (C1). Reference \cite{Vestues2022AgileStudy} provide further evidence for this challenge as they argue that federated governance can be unfeasible in specific situations and highlight the importance of central steering units to provide cross-domain insights. Furthermore, they raise concerns regarding the rightful access to data products \cite{Vestues2022AgileStudy}. Reference \cite{Podlesny2022CoK:Meshes} extend this argument and provide extensive analysis for privacy challenges with regard to the data mesh concept. Additionally, researcher describe agreement among domains and compliance issues as a ``moving target'' \cite{Joshi2021DataStudy}. 
In alignment with reference \cite{Vestues2022AgileStudy}, we identify the shift of responsibility as a main challenge (C2). The authors argue that domains lack competencies, motivation, and resources to provide data products \cite{Vestues2022AgileStudy}. Moreover, limited resources (C5), as well as acceptance and resistance issues (C6) are well-established challenges in the context of organizational and cultural change management; professionals and researchers should consider the rich body of literature to complement the findings of this work \cite[c.f.]{Oakland2007SuccessfulManagement, By2007OrganisationalReview, Joshi2021DataStudy, Vestues2022AgileStudy}.
As the interviews reveal, the adoption of modern architectures (MF5) can be associated with a low comprehension of the data mesh concept (C4), as adopters do not want to ''miss the boat''---instead of having deeply factual rooted motivation. This relationship finds little attention in recent publications, which is surprising as interviewees mention the challenge across multiple use cases. Neither does metadata quality (C3) in the specific context of data mesh. 

Addressing the aforementioned challenges, we formulate implementation strategies (IS1-6). First, to address C1 and C3, we argue that professionals should introduce a cross-domain steering unit (IS1). A recent study supports this claim. Reference \cite{Whyte2022DataPlatform} reasons that organizations should embrace decentralized and centralized capabilities for the \textit{next-generation data platform}. Furthermore, we argue that organizations need to empower domains and closely observe and track their progress (IS2) to address C2, C3, and C5. In line with reference \cite{Dehghani2022DataScale}, we describe the need to empower domains. Nevertheless, the great importance of observing data products as motivation for providers and for prioritization purposes is novel and needs further research. Furthermore, reference \cite{Dehghani2022DataScale} suggests an iterative approach for the adoption of the data mesh concept. We can confirm that approach after the analysis of our interview transcripts and provide further insights in IS3 on how to navigate organizational challenges (C5, C6).

Conscious adoption (IS4) is an important aspect when implementing a data mesh. Organizations should first go through a thorough assessment of their current data architecture before adopting data mesh in an organization---instead of a purely hype-related motivation \cite[c.f.]{Burgess2018DontHype}. We further suggest professionals favor small dedicated teams over a larger number of employees for the provision of data products (IS5). This implementation strategy addresses the need to prioritize data products using limited resources (C2, C5) \cite{Vestues2022AgileStudy}. Lastly, we argue that organizations should consider extending the data product owners' responsibilities or creating a data product steward role (IS6) to ensure data quality, compliance, fitness, and accessibility. In cases where organizations have a strong need to control data product access and overview use cases, this may be an interesting consideration. However, we acknowledge that IS6 somewhat conflicts with Dehghani's \cite{Dehghani2022DataScale} emphasis on the easy accessibility of data products throughout the organization and may create new bottlenecks or silos (MF1, MF3)---therefore, IS6 needs to be applied carefully. 

Across all interviews, interviewees state the positive impacts of their data mesh adoption effort. Our findings include improved accessibility (I1), speed (I2), data quality (I3), a reduction of redundancies (I4), a more data-driven organization (I5), an increase in trust (I6), reliability (I7), and data and analytics use cases (I8). The overall positive impacts highlight that organizations can reap the expected benefits even in the early stages of the adoption, confirming Dehghani's \cite{Dehghani2022DataScale} perspective on data mesh impacts.

We introduce two organizational archetypes embracing the data mesh concept: Startup \& scaleup (A1) and established organizations (A2). Professionals should assess implementation strategies based on their unique situations, for that purpose, A1 and A2 provide first steps to more specific guidelines.

\subsection{Contributions \& Implications}
\label{subsec:contributionsImplications}
This work enhances the empirical understanding of data mesh, addressing a vital gap in the existing scientific literature, which is currently limited and largely theoretical due to the novelty of the research area. With only a small number of independent, scholarly works exploring the practical aspects of data mesh, this paper provides vital insights that are both comprehensive and scientifically rigorous. 

We identify motivational factors (MF1-6) for organizations adopting data mesh, such as the need for agility, scalability, and improved data accessibility. Understanding these factors aids in justifying investment and effort in transitioning to data mesh architectures.

Further, this research identifies several challenges (C1-C6) that enable organizations contemplating the implementation of a data mesh to prepare and develop strategies to address these challenges proactively, thus mitigating risks that could potentially jeopardize the data mesh implementation.

Next, our outlined implementation strategies (IS1-6), tailored to address specific challenges, offer preliminary guidelines that can facilitate successful implementation and reduce associated risks.

Moreover, researching the impacts (I1-8) of adopting a data mesh is crucial as it provides empirical evidence on the benefits, helping organizations to assess the real-world effectiveness and scalability of this architecture enabling better strategic planning.

Last, archetypes based on organizational characteristics can guide companies in customizing their approach to data mesh adoption---highlighting more likely challenges to occur and corresponding implementation strategies.

The implications of this study are profound, particularly given the scarcity of independent research on the topic. For practitioners, this paper serves as a preliminary guide to understanding and implementing data mesh in a way that is both strategic and aligned with their specific organizational challenges. The insights it brings to addressing common pitfalls and leveraging effective strategies can significantly reduce the risk and enhance the success of data mesh initiatives.

This research also fills a gap for the academic community, providing a well-documented analysis of how data mesh is being implemented across different industries. It sets a foundation for future research to build upon, particularly in exploring how these strategies and challenges evolve as data mesh matures and becomes more widely adopted. 

Finally, this work serves as a valuable reference for both academics and practitioners in the landscape of (big) data management. It offers a clearer understanding of motivations, strategies, and impacts and transparently communicates these aspects while also highlighting potential challenges. In this way, it helps to position data mesh in the broader landscape of (big) data management.

\subsection{Limitations \& Future Research}
To answer our research question, we conduct 15 semi-structured expert interviews. 
The qualitative nature of our work results in limited quantitative validity. However, we justify the qualitative approach with the novelty of the research topic. 
Nevertheless, future research should investigate the findings on a quantitative level, e.g., using surveys.

With respect to our sampling method, we acknowledge a potential bias, recognizing that interviewees might frame their data mesh implementations as more successful than they actually are, especially as they aim to publicly position themselves as leaders in the domain of data mesh. To address this challenge, we have been transparent in communicating our focus solely on research activities. Moreover, we anonymize interviewee data to encourage them to speak freely about both the positive and negative aspects of their experiences within their organizations. This approach is designed to mitigate bias and ensure a more accurate and nuanced understanding of data mesh implementations.

Our main contribution are industry insights for professionals adopting a data mesh. However, we acknowledge that these practices are only applicable to a certain degree depending on the organizations' individual situation. Thus, professionals need to adjust their strategy accordingly---incorporating only relevant aspects.

The presented organizational archetypes are a first step towards creating more fine-grained guidelines for organizations based on specific characteristics. However, we acknowledge that the industry insights in general and the archetypes specifically lack quantitative evidence. This creates an enormous opportunity for future work, as this exploratory qualitative approach can be complemented with a quantitative study. 
In this context, researchers could investigate data mesh in small and medium-sized organizations to extend the framework of archetypes.

Moreover, future work should more deeply cover the technological realization of the data mesh concept. This could include designing technical data products in detail and integrating data warehouses, data lakes, or blob storage to realize the data mesh concept. In addition, possible data mesh topologies and considerations for ideal data mesh node sizes can be explored in more detail. 

In summary, a promising field of research lies ahead.

\section{Conclusion}
\label{sec:conlusion}
This paper provides a comprehensive exploration of the adoption of the data mesh concept across various industries based on insights derived from 15 semi-structured expert interviews. The study addresses a gap in the existing literature by offering practical, empirical insights into the motivations, challenges, implementation strategies, and impacts of data mesh implementations, which have been largely theoretical to this point due to the novelty of the concept.

The findings reveal that motivations for adopting a data mesh include the desire to reduce bottlenecks, leverage domain knowledge, improve data ownership, and break down data silos, all aimed at enhancing data accessibility and quality. These motivations align well with the theoretical benefits posited by Dehghani's foundational framework on data mesh \cite{Dehghani2022DataScale}.

However, the transition to a data mesh architecture is not without challenges. These include the complexities of shifting to federated governance, managing the responsibilities that come with decentralized data ownership, ensuring high-quality metadata, and addressing the organizational resistance that can accompany significant changes in data management practices. To overcome these obstacles, the study proposes several implementation strategies, such as establishing cross-domain units, empowering and closely observing domain efforts, securing quick wins, promoting conscious adoption, enforcing dedicated ownership, and considering the role of data stewards.

The impacts observed from early implementations of data mesh are promising, including improved accessibility and speed of data access, enhanced data quality, reduction of redundancies, and an overall progression towards a more data-driven organization. These impacts confirm the potential of data mesh to improve organizational data management practices significantly.

The study also identifies two preliminary organizational archetypes—startups \& scaleups and established organizations—that benefit from tailored approaches to data mesh implementation. This differentiation helps in understanding how data mesh can be adapted to fit the specific needs and characteristics of different types of organizations.

By providing a detailed analysis of real-world experiences with data mesh, this paper contributes to both academic research and practical applications in data management. It lays the groundwork for further studies and helps organizations better prepare for the challenges and opportunities that come with adopting a data mesh architecture. Future research should continue to explore these themes in a complementing quantitative fashion as data mesh matures and its adoption becomes more widespread.

\bibliographystyle{ACM-Reference-Format}
\bibliography{references3}

\appendix

\end{document}